\documentclass[runningheads]{llncs}
\usepackage{graphicx}
\usepackage{amsmath,amssymb} 
\usepackage{color}
\usepackage[width=122mm,left=12mm,paperwidth=146mm,height=193mm,top=12mm,paperheight=217mm]{geometry}
\usepackage[pagebackref=true,breaklinks=true,letterpaper=true,colorlinks,bookmarks=false]{hyperref}

\usepackage{multirow,bm}

\newcommand{\R}{\mathbb{R}}
\newcommand{\logs}{\textrm{logs}}
\newcommand{\Proj}{\textrm{Proj}}
\newcommand{\Retr}{\textrm{Retr}}
\newcommand{\etal}{{et al.~}}

\begin{document}
\pagestyle{headings}
\mainmatter
\def\ECCV18SubNumber{2124}  

\title{Statistical transformer networks: learning shape and appearance models via self supervision}

\titlerunning{Statistical transformer networks}

\authorrunning{A. Bas and W. A. P. Smith}

\author{Anil Bas and William A. P. Smith}
\institute{Department of Computer Science, University of York\\{\tt\small \{ab1792,william.smith\}@york.ac.uk}}

\maketitle

\begin{abstract}
We generalise Spatial Transformer Networks (STN) by replacing the parametric transformation of a fixed, regular sampling grid with a deformable, statistical shape model which is itself learnt. We call this a Statistical Transformer Network (StaTN). By training a network containing a StaTN end-to-end for a particular task, the network learns the optimal nonrigid alignment of the input data for the task. Moreover, the statistical shape model is learnt with no direct supervision (such as landmarks) and can be reused for other tasks. Besides training for a specific task, we also show that a StaTN can learn a shape model using generic loss functions. This includes a loss inspired by the minimum description length principle in which an appearance model is also learnt from scratch. In this configuration, our model learns an active appearance model and a means to fit the model from scratch with no supervision at all, even identity labels.
\end{abstract}

\section{Introduction}

Establishing correspondence between images of objects from the same class is a fundamental task in computer vision. It enables appearance to be disentangled from the effects of pose and shape deformation, simplifying the task of comparing objects. While Convolutional Neural Networks (CNNs) are explicitly invariant to small, local translations (through pooling), they have no obvious mechanism for establishing dense, possibly nonrigid correspondence. 

Statistical Transformer Networks (STNs) \cite{jaderberg2015spatial} introduce an explicit geometric transformation into CNNs. Rather than rely on generic layers within a CNN to learn invariance to various kinds of spatial transformation, an STN includes expert layers that predict and apply a parametric transformation to an input feature map. By including an STN as a component within a CNN, the network can learn its own notion of alignment that is optimal for the task that it is learning to solve. 

The components of an STN are: 1.~a localiser network (a CNN that learns to regress transformation parameters from an input feature map), 2.~a grid generator (that generates a grid of sample points from the transformation parameters) and 3.~a differentiable resampler (that uses bilinear interpolation to resample the input feature map at the sample grid locations).

In this paper we address an important drawback of the original STN. The geometric transformation model used by the STN must be hand picked and remains fixed. We propose to replace the hand picked transformation model by a statistical shape model that is itself learnt. We call this a Statistical Transformer Network (StaTN). The power of statistical shape and appearance models is well known, for example Active Appearance Models \cite{cootes2001active} in 2D and 3D Morphable Models \cite{blanz1999morphable} in 3D. These models can be built from a few hundred or thousand samples and then deployed to solve problems ranging from tracking to recognition to synthesis. Usually, constructing such models requires hand labelling of landmark points so that correspondence can be established between training samples. Then the variability in shape and appearance is learnt, typically using Principal Components Analysis (PCA). 

A StaTN learns such a statistical shape model (and optionally a statistical appearance model) with no landmark supervision and also learns to fit the model. Hence, a StaTN learns an explicit representation of a particular object class in an interpretable way (the parameters of the statistical model can be explicitly accessed and understood).

We make the following novel contributions:
\begin{itemize}
    \item We show how to incorporate a 2D statistical shape model in a spatial transformer network.
    \item The mean and principal components of a statistical model are subject to constraints (e.g.~orthogonality of principal components). We show how these can be enforced by incorporating manifold gradient descent into backpropagation.
    \item We introduce generic losses that can be used to train a StaTN, including a statistical appearance model, with no supervision at all (i.e. not even identity labels for computing a classification loss).
\end{itemize}

\begin{figure}[t]
\noindent\resizebox{\textwidth}{!}{
\includegraphics[trim={1cm 16cm 1cm 1cm},clip]{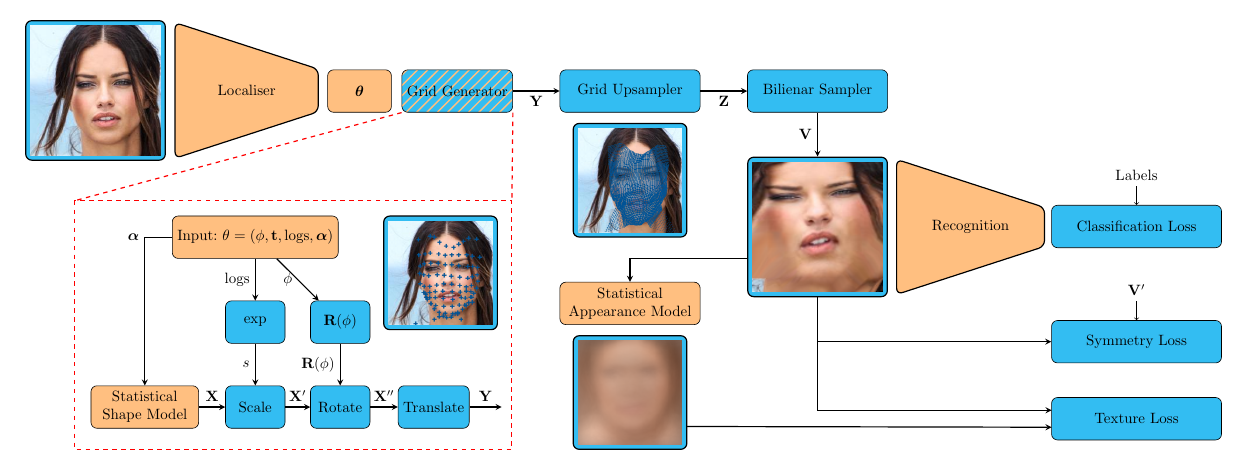}}
\caption{An overview of the statistical transformer network. Learnable components of the network are shown in orange. }
\label{fig:example}
\end{figure}

\section{Related Work}

The general problem of alignment has been approached from a learning perspective previously \cite{huang2012learning}. The specific question of whether CNNs learn a notion of correspondence implicitly has been considered by Long \etal \cite{long2014convnets}. More generally, morphable models (models of the warps between samples from an object class in dense correspondence) have been learnt directly from in the wild image data \cite{cashman2013shape,antonakos2014automatic}, though note that the process of fitting these models is tackled separately (using handcrafted algorithms).


\paragraph{Geometry Matching and Warping.}

Kanazawa \etal \cite{kanazawa2016warpnet} proposed a weakly supervised architecture that learns correspondences by shape deformation in a fine-grained dataset. Rocco \etal \cite{rocco2017convolutional} propose a neural network for geometric matching between two images by estimating an affine transformation. However, their procedure strongly relies on fully supervised training. Thelwis \etal \cite{thewlis2017unsupervised} followed an unsupervised approach to establish correspondences between different object instances in categories such as human and cat faces. This is later extended from a sparse set of landmarks to a dense model \cite{thewlis2017unsupervised2}.


\paragraph{Face Analysis.}

Zhu \etal \cite{zhu2014recover} recover the canonical frontal view of face images using deep neural networks while Dig \etal\cite{ding2015multi} handle the full range of pose variations in a multi-task learning fashion. Hassner \etal \cite{hassner2015effective} and Chang \etal \cite{chang2017faceposenet} estimate frontalisation by using a single 3D reference surface. These four studies reported that the existing face recognition methods can be improved with adopting their approach as normalisation and pre-processing. Zhu \etal \cite{zhu2015high} present a pose and expression normalisation method by applying 3D model fitting and texture warping which later extended to deep convolutional network \cite{zhu2016face}.
Jourabloo and Liu \cite{jourabloo2016large} fit a 3D model to detected landmarks and train a CNN to directly regress the fitted pose and shape parameters.
Ranjan \etal \cite{ranjan2017hyperface,ranjan2017all} proposed a multi-task architecture that simultaneously predicts landmark locations, estimates pose and identifies a face.


\paragraph{Transformer networks.}

Since the introduction of the Spatial Transformer Network \cite{jaderberg2015spatial}, there are many studies that presented the module in a similar manner from correspondence estimation \cite{choy2016universal} to image segmentation \cite{li2017dense} to radio communication \cite{o2016radio}.
Wu \etal \cite{wu2017recursive} apply recursive STNs in the context of face recognition.
Yu \etal \cite{yu2016deep} present the Deep Deformation Network to estimate landmarks to which local deformation is applied using thin-plate spline transformation as part of their point transformer network.
Dai \etal \cite{dai2017deformable} introduce the Deformable Convolutional Networks based on the idea of augmenting the spatial sampling locations in the modules with additional offsets and then learning the offsets from target tasks without additional supervision.
Bas \etal \cite{bas20173D} propose an architecture based on a purely geometric approach by incorporating a 3D Morphable Model into a Spatial Transformer Network to interpret and normalise pose changes and self-occlusions.
Zhong \etal \cite{zhong2017towards} implement the spatial transformer module for alignment learning in end-to-end face recognition.

\section{Statistical transformer networks}

In this section we explain how to incorporate a learnable statistical shape model into a spatial transformer network. We explain how each component of a conventional STN must be modified. An overview of our proposed StaTN is shown in Figure \ref{fig:example}.

\subsection{Localiser network}

The localiser network is a black box CNN that takes an image (or, more generally, a feature map) as input and regresses a semantically meaningful vector of parameters $\theta$:
\begin{equation}
    \theta = (\underbrace{\phi,\mathbf{t},\logs}_{\textrm{rigid pose}},\underbrace{{\bm \alpha}}_{\textrm{shape}}).
\end{equation}
Here, $\mathbf{t}\in\R^2$ is a 2D translation and $\phi\in\R$ is a rotation angle. Since scale must be positive, we estimate log scale ($\logs$) and later pass this through an exponentiation layer, ensuring that the estimated scale, $s$, is positive. The shape parameters ${\bm \alpha}\in\R^D$ are the weights of the principal components of the statistical shape model described below.

The architecture of the localiser network is not critical. For all our experiments, we use a very simple architecture comprising six blocks of convolution, ReLU and pooling, followed by a fully connected layer with 1024 units followed by the final regression layer implemented as a fully connected layer with $D+4$ units.

\subsection{Grid generator}

The purpose of the grid generator is to compute a sampling coordinate $(x_i^s,y_i^s)$ for each corresponding point $(x_i^t,y_i^t)$ in the regular output grid from the transformation parameters provided by the localiser network. The output grid comprises $M=H^{\prime}W^{\prime}$ points, regularly sampled over $-1\dots 1$ with height $H^{\prime}$ and width $W^{\prime}$. Our grid generator begins by generating a shape from a linear shape model (which is learnt as part of the StaTN training), then a rigid transformation is applied to this before it is finally upsampled to a high resolution sampling grid.

\paragraph{Linear shape model.}
A linear shape model is an orthonormal basis enabling compact representation of a class of shapes. Specifically, a shape comprised of $N$ 2D vertices, $\mathbf{x}\in\R^{2N}$, is written as a sum of a mean shape $\mathbf{b}\in\R^{2N}$ and a linear combination of a set of $D$ orthonormal bases $\mathbf{F}\in\R^{2N\times D}$:
\begin{equation}
    \mathbf{x} = \mathbf{F}\alpha + \mathbf{b},\label{eqn:linshape}
\end{equation}
where $\alpha\in\R^D$ is a vector of shape parameters and $\mathbf{F}^T\mathbf{F}=\mathbf{I}_D$. Typically, such models are built statistically by labelling a set of training images and using PCA to extract the mean and basis vectors. Instead, here we will learn the model in an unsupervised manner simultaneously with learning to fit the model.

Note that the linear model in \eqref{eqn:linshape} can be interpreted as a fully connected layer of a CNN (or equivalently, a special case of a convolution layer) in which the orthonormal basis plays the role of the filters, the mean shape plays the role of the biases and the parameter vector plays the role of the input feature map. To make this explicit, we rewrite each component of the model in tensor form such that the output shape is ${\cal X}\in\R^{1\times 1\times 2N}$ and the model is given by the orthonormal basis ${\cal F}\in\R^{D\times 1\times 1\times 2N}$ and the parameter vector ${\cal \alpha}\in\R^{D\times 1 \times 1}$. In this form, the familiar definition of a convolution operation yields the same model as \eqref{eqn:linshape}:
\begin{equation}
    {\cal X}_{i^{\prime},j^{\prime},k^{\prime}}=\mathbf{b}_{k^{\prime}} + \sum_{i,j,k} {\cal F}_{i,j,k,k^{\prime}}\alpha_{i+i^{\prime},j+j^{\prime},k}.
\end{equation}
In practice, we implement the linear shape model as a convolution layer and learn the filters and biases as normal. We initialise the biases (mean shape) as a regular, square grid and the filters (principal components) as a random orthonormal matrix.

Subsequently, it is notationally convenient to rewrite the output shape and mean shape in matrix form as $\mathbf{X}\in\R^{2\times N}$ and  $\mathbf{B}\in\R^{2\times N}$.

\paragraph{Scaling Layer.}

The log scale estimated by the localiser is first transformed to scale by an exponentiation layer:
\begin{equation*}
    s(\logs)=\exp(\logs),\quad \frac{\partial s}{\partial \logs}=\exp(\logs).
\end{equation*}
Then, the 2D points $\mathbf{X}\in\R^{2\times N}$ are scaled:
\begin{equation*}
    \mathbf{X}^{\prime}(s,\mathbf{X})=s\mathbf{X},\quad \frac{\partial X^{\prime}_{i,j}}{\partial s}=X_{i,j},\quad \frac{\partial X^{\prime}_{i,j}}{\partial X_{i,j}}=s
\end{equation*}

\paragraph{2D Rotation Matrix.}
This layer outputs a 2D rotation matrix as a function of a rotation angle $\mathbf{R}:\R\mapsto\R^{2\times 2}$:
\begin{equation}
    \mathbf{R}(\phi) = \begin{bmatrix}\cos\phi & -\sin\phi \\ \sin\phi & \cos\phi \end{bmatrix},\quad
    \frac{\partial \mathbf{R}}{\partial \phi} = \begin{bmatrix}-\sin\phi & -\cos\phi \\ \cos\phi & -\sin\phi \end{bmatrix}.
\end{equation}

\paragraph{2D Rotation Layer.}

The rotation layer takes as input a rotation matrix $\mathbf{R}$ and $N$ 2D points $\mathbf{X}^{\prime}\in\R^{2\times N}$ and applies the rotation:
\begin{gather*}
    \mathbf{X}^{\prime\prime}(\mathbf{R},\mathbf{X}^{\prime})=\mathbf{RX}^{\prime}\\
    \frac{\partial X^{\prime\prime}_{i,j}}{\partial R_{i,k}}=X^{\prime}_{k,j},\quad \frac{\partial X^{\prime\prime}_{i,j}}{\partial X^{\prime}_{k,j}}=R_{i,k},\quad i,k\in\{1,2\}, j\in\{1,\dots,N\}.
\end{gather*}

\paragraph{Translation Layer.}
Finally, the 2D sample points are generated by adding a 2D translation $\mathbf{t}\in\R^2$ to each of the scaled points:
\begin{equation*}
    \mathbf{Y}(\mathbf{t},\mathbf{X}^{\prime\prime})=\mathbf{X}^{\prime\prime}+\mathbf{1}_{N}\otimes \mathbf{t},\quad \frac{\partial Y_{i,j}}{\partial t_i}=1,\quad \frac{\partial Y_{i,j}}{\partial X^{\prime\prime}_{i,j}}=1,
\end{equation*}
where $\mathbf{1}_{N}$ is the row vector of length $N$ containing ones and $\otimes$ is the Kronecker product.

\subsection{Grid upsampler}

The resolution at which we wish to resample the image may be higher than the resolution at which we wish to statistically model shape. E.g. in our experiments, our statistical shape model comprises $N = 10\times 10$ grid vertices whereas our resampled images comprise $M = 112\times 112$ pixels, i.e. two orders of magnitude more. This keeps the dimensionality of the statistical model (that must be learnt from data) down, whilst still allowing sufficient detail to be sampled from the input images. To achieve this, we precompute the barycentric weights of each high resolution output grid point in the low resolution output grid. We then use these weights to compute sample locations for every high resolution point, $\mathbf{Z}\in\R^{2\times M}$, from the computed low resolution sample grid points. In other words, we perform a linear interpolation of the low resolution sample grid. In practice, this can be written as:
\begin{equation}
    \mathbf{Z}(\mathbf{Y}) = \mathbf{YW},\quad  \frac{\partial Z_{i,j}}{\partial Y_{i,k}}=W_{k,j},\quad i\in\{1,2\},\, j\in\{1,\dots,M\},\, k\in\{1,\dots,N\}.
\end{equation}
where $\mathbf{W}\in\R^{N\times M}$ is constant, sparse (each row contains three non-zero values) and each row sums to one: $\mathbf{W1}_M=\mathbf{1}_N$.
The sample points for each point in the output grid are given by $(x_i^s,y_i^s)=(Z_{1,i},Z_{2,i})$. See Figure \ref{fig:example} for a visualisation of the low and high resolution sampling grids overlaid on an input image.

\subsection{Bilinear sampling}
We use bilinear sampling, exactly as in the original STN such that the re-sampled image $V_i^c$ at location $(x_i^t,y_i^t)$ in colour channel $c$ is given by:
\begin{equation*}
    V_i^c = \sum_{j=1}^H\sum_{k=1}^W I_{jk}^c \max(0,1-|x_i^s-k|)\max(0,1-|y_i^s-j|)
\end{equation*}
where $I_{jk}^c$ is the value in the input image at pixel $(j,k)$ in colour channel $c$. $I$ has height $H$ and width $W$. This bilinear sampling is differentiable (see \cite{jaderberg2015spatial} for derivatives) and so the loss can be backpropagated through the sampler and back into the grid generator.

\section{Backpropagation with Manifold Gradient Descent}\label{sec:manopt}

In a StaTN, some learnable parameters are subject to constraints. If, during backpropagation, an unconstrained step in the direction of the negative gradient of the loss function is taken, then these parameters will no longer satisfy the constraints. In this section, we show how manifold gradient descent can be used to ensure constraints on learnable parameters remain satisfied during training. 

\subsection{Constrained parameters}

In our network, the shape model is subject to such constraints and hence requires special treatment during training. First, the shape basis is required to be orthonormal, i.e. that $\mathbf{F}^T\mathbf{F}=\mathbf{I}_D$. Second, we require that the mean shape is centred, i.e. that $\mathbf{B1}_N=\mathbf{I}_2$. Otherwise there is an ambiguity between the translation estimated by the localiser, $\mathbf{t}$, and the centering of the mean (i.e. the same shape can be obtained by translating the mean or translating the output shape from our model). Without constraint, this gives SGD redundant search directions during training.

Both of these constraints can be encoded by viewing the parameters as belonging to a Riemannian manifold and using manifold optimisation for these parameters during training. This idea is not new and has been considered, for example, by Harandi and Fernando \cite{harandi2016generalized}. Here, we show how to use manifold optimisation for the two model parameters in our STN that are subject to constraints.

\subsection{Manifold gradient descent}

Suppose $M\subset \R^n$ is a Riemannian manifold embedded in $\R^n$ and $f:\R^n\mapsto\R$ a cost function on $\R^n$. If $\mathbf{x}\in\R^n$ is some learnable parameter then $-\nabla f(\mathbf{x})$ is a (Euclidean) descent direction for $\mathbf{x}$. In practice, this Euclidean gradient would be provided by backpropagation. Usually, some variation of stochastic gradient descent is used to reduce the loss by taking a step in the negative gradient direction. However, if our learnable parameters are subject to constraints then taking a step in the unconstrained gradient direction will lead to parameters that do not satisfy the constraints. 

Manifold optimisation relies on two operations: {\it orthogonal projection} from the ambient space to the tangent space of the manifold and {\it retraction} to transform from the tangent space onto the manifold. The Euclidean gradient computed via backpropagation is first projected to the tangent space, then a retraction is applied to this tangent vector, giving a new point on the manifold. Note that the geometric exponential map is a particular kind of retraction but often we can use alternatives that are cheaper to compute.


\begin{figure}[!t] \centering
\noindent\resizebox{0.65\textwidth}{!}{
\includegraphics[trim={1cm 15cm 1cm 1cm},clip]{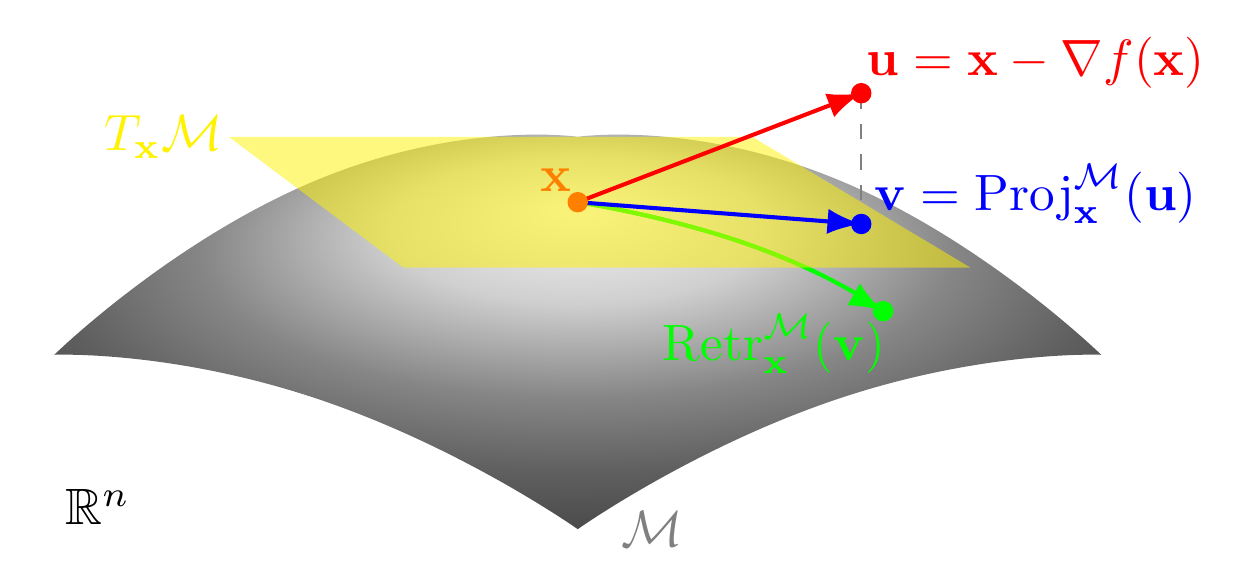}}
\caption{Manifold optimisation: the Euclidean descent direction $-\nabla f(\mathbf{x})$ is transformed to the tangent plane at ${\bf x}$, $T_{\mathbf{x}}{\cal M}$, via orthogonal projection and then to the manifold ${\cal M}\subset\R^n$ via a retraction.}
\label{fig:manifoldopt}
\end{figure}

\paragraph{Centred matrices manifold.} The mean shape must lie on the manifold of centred matrices $C_{m,n}=\{\mathbf{X}\in\R^{m\times n} | \mathbf{X1}_n=\mathbf{0}_m \}$. Specifically, $\mathbf{B}\in C_{2,N}$. Without this constraint there is a translational ambiguity between the translation vector $\mathbf{t}$ and the mean shape. Projection and retraction on this manifold are particularly simple. The orthogonal projection $\Proj_{\mathbf{X}}^{C_{m,n}}:\R^{m\times n}\mapsto T_{\mathbf{X}}C_{m,n}$ of a displacement $\mathbf{U}\in \R^{m\times n}$ in the ambient space onto the tangent space $T_{\mathbf{X}}C_{m,n}$ at $\mathbf{X}$ is obtained simply by centering $\mathbf{U}$:
\begin{equation}
\Proj_{\mathbf{X}}^{C_{m,n}}(\mathbf{U}) = \mathbf{U}-\mathbf{U1}_n.
\end{equation}
The retraction $\Retr_{\mathbf{X}}^{C_{m,n}}:T_{\mathbf{X}}C_{m,n}\mapsto C_{m,n}$ of a tangent vector $\mathbf{V}\in T_{\mathbf{X}}C_{m,n}$ is simply:
\begin{equation}
\Retr_{\mathbf{X}}^{C_{m,n}}(\mathbf{V}) = \mathbf{X}+\mathbf{V}.
\end{equation}
So, we initialise with a centred shape then, when updating the mean shape during SGD, we simply centre the gradient provided by backpropagation before adding it to the current mean shape.

\paragraph{Stiefel manifold.} The orthonormal shape basis must lie on the Stiefel manifold $V_k(\R^n)=\{\mathbf{X}\in\R^{n\times k} | \mathbf{X}^T\mathbf{X}=\mathbf{I}_k\}$. This is the manifold of $k$ dimensional orthonormal bases in $\R^n$. Specifically, $\mathbf{F}\in V_D(\R^{2N})$. The orthogonal projection $\Proj_{\mathbf{X}}^{V_k(\R^n)}:\R^{n\times k}\mapsto T_{\mathbf{X}}V_k(\R^n)$ of a displacement $\mathbf{U}\in \R^{n\times k}$ in the ambient space onto the tangent space $T_{\mathbf{X}}V_k(\R^n)$ at $\mathbf{X}$ is given by:
\begin{equation}
    \Proj_{\mathbf{X}}^{V_k(\R^n)}(\mathbf{U}) = \mathbf{U}-\mathbf{X}\,\textrm{sym}(\mathbf{X}^T\mathbf{U}),
\end{equation}
where $\textrm{sym}(\mathbf{M})=0.5(\mathbf{M}+\mathbf{M}^T)$. A retraction $\Retr_{\mathbf{X}}^{V_k(\R^n)}:T_{\mathbf{X}}V_k(\R^n)\mapsto V_k(\R^n)$ of a tangent vector $\mathbf{V}\in T_{\mathbf{X}}V_k(\R^n)$ can be obtained by finding the closest orthogonal matrix to $\mathbf{V}$:
\begin{equation}
    \Retr_{\mathbf{X}}^{V_k(\R^n)}(\mathbf{V}) = \mathbf{U},
\end{equation}
where $\mathbf{V} = \mathbf{UP}$ is the polar decomposition of $\mathbf{V}$.

\subsection{Implementation}

In practice, we make a small modification to the implementation of backpropagation. Where layer parameters are updated, we test whether the layer is one with constraints. If it is, we apply projection and retraction before the updates are added to the parameters.

\section{Losses for training a StaTN}

As with the original STN, a StaTN can be used as a component within a larger network that is trained end-to-end. In this section, we consider some different ways that this can be achieved and design loss functions that help the StaTN learn a meaningful statistical model.

\subsection{Learning by task}

The most obvious way to use a StaTN is as part of a network that is trained to solve a task such as recognition or classification. Here, the StaTN acts to normalise the effects of pose and shape, making the subsequent task easier to solve. Concretely, the output of the StaTN (i.e. the resampled image) is fed to a classification network with, for example, its own softmax loss (see Figure \ref{fig:example}). This loss is propagated back through the classification network, through the resampler and into the statistical shape model and the localisation networks.

In this setting, the StaTN will learn a notion of correspondence that is optimal for the task being solved. This may not coincide with intuitive notions of correspondence, nor will attention necessarily focus only on the object of interest. For example, if training for face recognition, a StaTN may learn that there is important contextual information in clothing or background and so the statistical model (and hence sample grid) may not attend only to the face.

In our experiments, we use a softmax classification loss, $\ell_{\textrm{class}}$, in the context of a face classification task.

\subsection{Appearance model with minimum description length}

We now propose a loss that can be used to train a StaTN in a much more general setting. Consider that we have an image collection containing images of a particular object class but no further information, i.e. we do not even have identity labels for each image. The minimum description length principle \cite{davies2002minimum} asserts that the correct correspondence is the one that leads to the best compression of the data. We apply this principle by learning a statistical appearance model of the images obtained by a StaTN resampling and then measuring a loss as the reconstruction error of the images. This loss will be minimised by the StaTN learning to establish correspondence that leads to the most compressible appearance. Intuitively, in the image collection, the StaTN then searches for objects with the most redundant appearance. 

Specifically, we learn a linear statistical appearance model of the resampled images $\mathbf{V}\in\R^{H^{\prime}W^{\prime}\times C}$, where $C$ is the number of colour channels (usually the StaTN and hence the appearance model will be applied to RGB image input and so $C=3$, however in general this approach could be applied to feature maps with any number of channels). We use the same linear model as in \eqref{eqn:linshape}, where $\mathbf{b}$ is the mean texture and $\mathbf{F}$ the texture principal components. The projection of an input image onto the model is given by:
\begin{equation}
    \mathbf{w} = \mathbf{F}\mathbf{F}^T(\textrm{vec}(\mathbf{V})-\mathbf{b})+\mathbf{b}.
\end{equation}
Note that this is just a linear autoencoder. A more complex, nonlinear autoencoder could be used here for the texture model but it has been shown many times previously that a linear model is an efficient representation for the appearance of many object classes \cite{cootes2001active}. The texture loss is then given by the squared Euclidean distance between the source and reconstructed textures:
\begin{equation}
    \ell_{\textrm{tex}} = \|\mathbf{w}-\textrm{vec}(\mathbf{V})\|^2.
\end{equation}
Note that the principal components of the texture model are subject to the same orthogonality constraint as the shape model, i.e. they lie on the Stiefel manifold. Hence, we use the same manifold optimisation strategy for these parameters as in Section \ref{sec:manopt}. The mean texture does not need constraining since there is no texture translation to cause an ambiguity.

Note that, when trained in this way, a StaTN is effectively learning an Active Appearance Model \cite{cootes2001active} and the means to fit the model to an image with no supervision.

\subsection{Regularisation}

Besides the above two losses, we may wish to regularise the process of training a StaTN such that the obtained shape and appearance models exhibit desirable properties.

\paragraph{Symmetry loss.}
Many natural and man-made objects exhibit bilateral symmetry. Usually, statistical shape and appearance models would be symmetric by construction since the chosen landmarks would be symmetric. However, we do not use landmarks and neither the classification loss nor the texture loss require this to be the case. To encourage a symmetric model we penalise asymmetry, measured as the difference between a sampled image and its reflection:
\begin{equation}
    \ell_{\textrm{sym}} = \sum_{i=1}^M \sum_{c=1}^3 (V_i^c - V_{\textrm{sym}(i)}^c)^2,
\end{equation}
where $V_{\textrm{sym}(i)}^c$ is the value in the resampled image at location $(W^{\prime}+1-x_i^t,y_i^t)$.
This ignores the effect of illumination (which may introduce asymmetries in appearance) but is still a useful regulariser when averaged over batches.

\paragraph{Area loss.}
When training without a classification loss, i.e. using only the texture loss, a trivial solution is to collapse the grid to a single pixel. This makes the appearance constant and hence compressible. To avoid this we propose a second regularisation. In a triangulation of our sample grid, we would like the area of the triangles to be preserved (i.e. not collapase to zero). More generally, we would like our shape model to be diffeomorphic, i.e. avoid triangles folding over themselves. Hence, for a sample grid we compute the signed area, $a_t$, for each triangle $t$ and penalise areas close to zero or that are negative (i.e. have flipped) as follows:
\begin{equation}
    \ell_{\textrm{area}} = \sum_t \max(0,\exp(-a_t)-k),
\end{equation}
where $0< k\leq 1$ is a constant which determines how small a triangle must be before the penalty is applied. $k=1$ means only negative areas are penalised. $k$ close to zero means even large triangle areas are penalised. We use a value of $k=0.99$ in our experiments.

\subsection{Hybrid loss}

In our experiments, we use a hybrid loss function comprising a weighted sum of the four losses (where a loss is switched off by setting the corresponding weight to zero):
\begin{equation}
    \ell = w_{\textrm{class}}\ell_{\textrm{class}} + w_{\textrm{tex}}\ell_{\textrm{tex}} + w_{\textrm{sym}}\ell_{\textrm{sym}} + w_{\textrm{area}}\ell_{\textrm{area}}.
\end{equation}

\begin{figure}[!t]
\centering
\noindent\resizebox{.65\textwidth}{!}{
\includegraphics[clip]{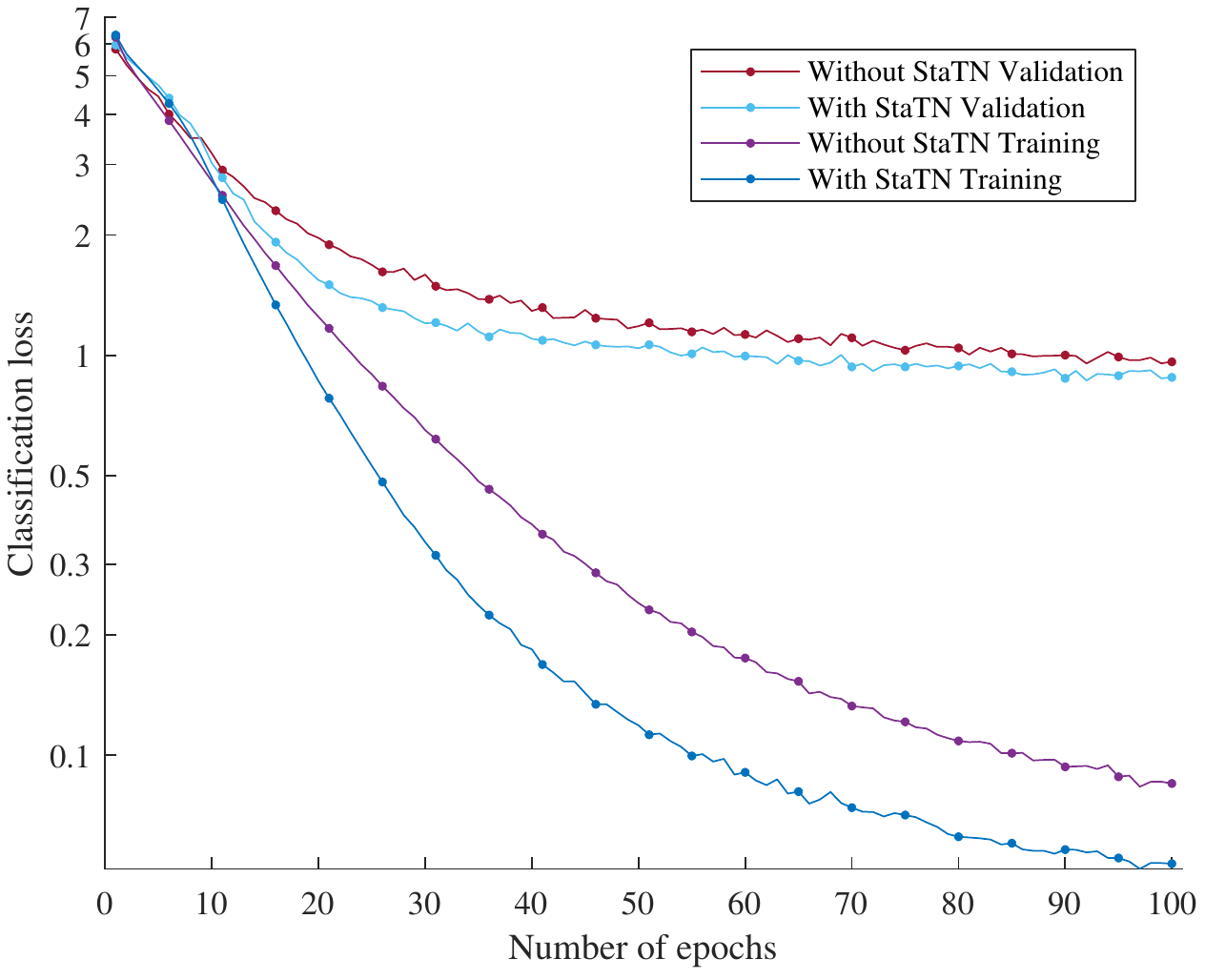}
}
\caption{Training and validation curves.}
\label{fig:curve}
\end{figure}

\begin{figure}[!t]
\centering
\noindent\resizebox{\textwidth}{!}{
\begin{tabular}{ccccccc}

\includegraphics[clip=true]{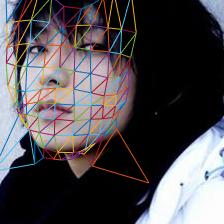}&
\includegraphics[clip=true]{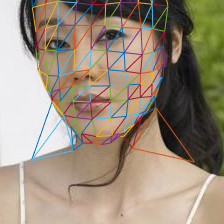}&
\includegraphics[clip=true]{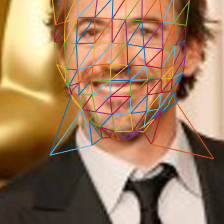}&
\includegraphics[clip=true]{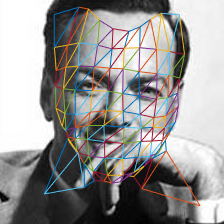}&
\includegraphics[clip=true]{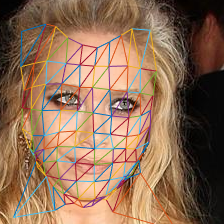}&
\includegraphics[clip=true]{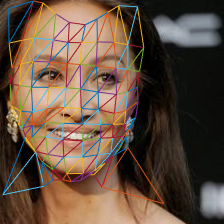}&
\includegraphics[clip=true]{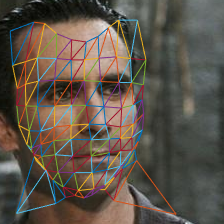}\\
\includegraphics[clip=true]{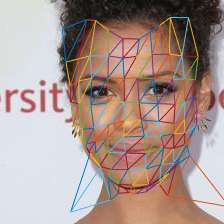}&
\includegraphics[clip=true]{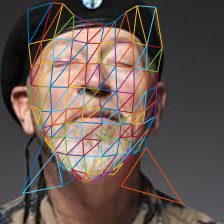}&
\includegraphics[clip=true]{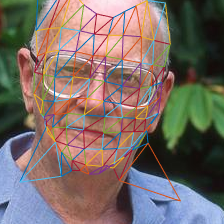}&
\includegraphics[clip=true]{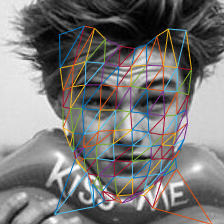}&
\includegraphics[clip=true]{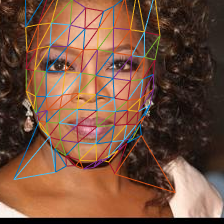}&
\includegraphics[clip=true]{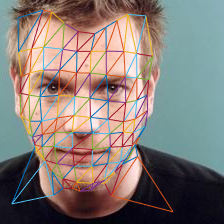}&
\includegraphics[clip=true]{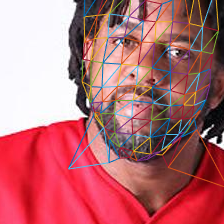}\\
\includegraphics[clip=true]{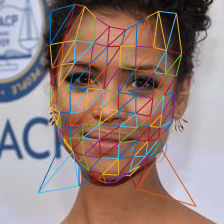}&
\includegraphics[clip=true]{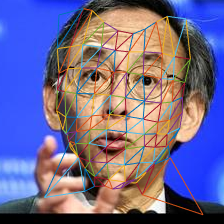}&
\includegraphics[clip=true]{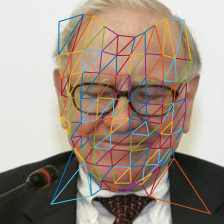}&
\includegraphics[clip=true]{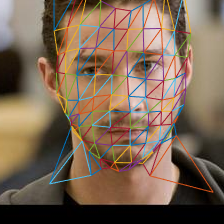}&
\includegraphics[clip=true]{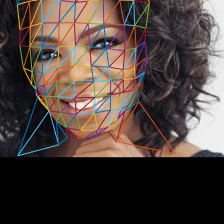}&
\includegraphics[clip=true]{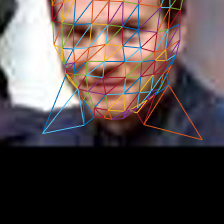}&
\includegraphics[clip=true]{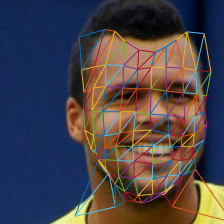}

\end{tabular}
}
\caption{Qualitative StaTN fitting results. We show a triangulation of the low resolution sample grid predicted by the grid generator. The deformable grid and fitting process have been learnt from scratch in an end-to-end trained face classification network with no landmark supervision.}
\label{fig:fitting}
\end{figure}

\begin{figure}[!t] \centering
\noindent\resizebox{\textwidth}{!}{

\begin{tabular}{cccc}
\hline
\multirow{2}{*}{\bfseries Mean Shape} & 
\multicolumn{3}{c}{\bfseries Shape Components}\\ \cline{2-4}
& 1st & 2nd & 3rd \\ \hline\\[-0.5em]
\includegraphics[height=2cm, clip=true]{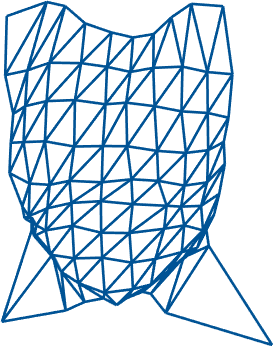}&
\includegraphics[height=2cm, clip=true]{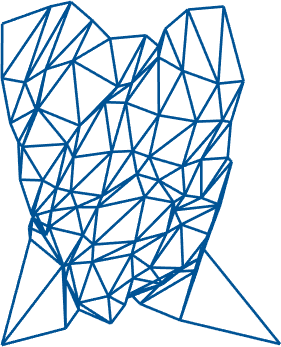}&
\includegraphics[height=2cm, clip=true]{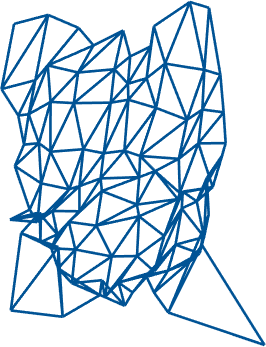}&
\includegraphics[height=2cm, clip=true]{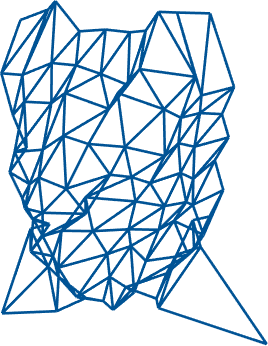}\\
&
\includegraphics[height=2cm, clip=true]{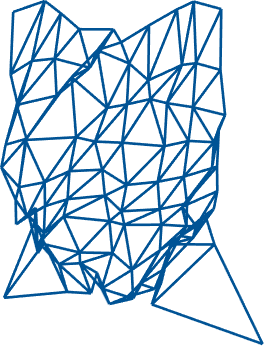}&
\includegraphics[height=2cm, clip=true]{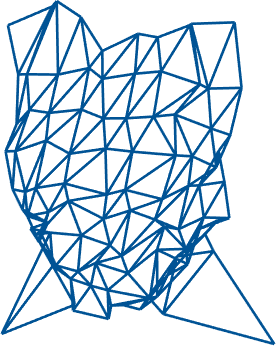}&
\includegraphics[height=2cm, clip=true]{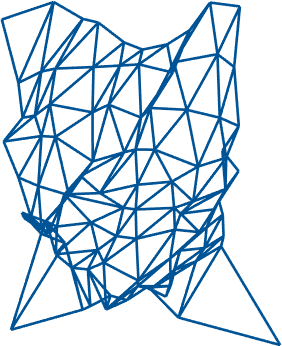}
\end{tabular}

\quad

\begin{tabular}{cccc}
\hline
\multirow{2}{*}{\bfseries Mean Texture} & 
\multicolumn{3}{c}{\bfseries Texture Components}\\ \cline{2-4}
& 1st & 2nd & 3rd \\ \hline\\[-0.5em]
\includegraphics[height=2cm, clip=true]{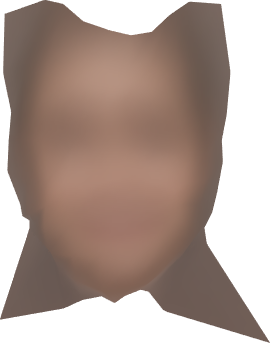}&
\includegraphics[height=2cm, clip=true]{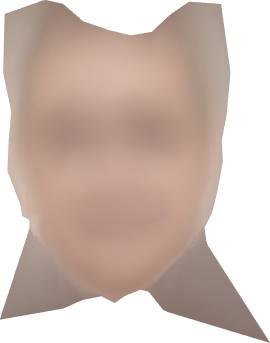}&
\includegraphics[height=2cm, clip=true]{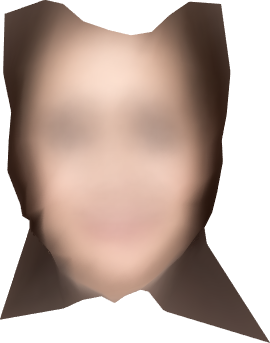}&
\includegraphics[height=2cm, clip=true]{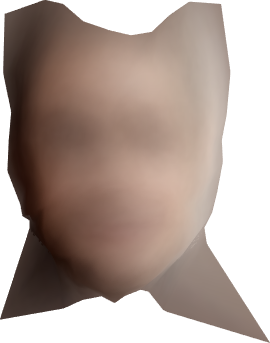}\\ 
&
\includegraphics[height=2cm, clip=true]{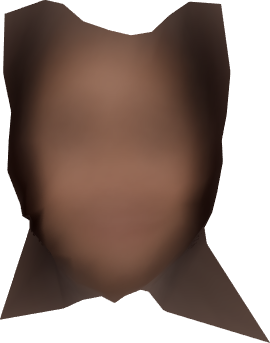}&
\includegraphics[height=2cm, clip=true]{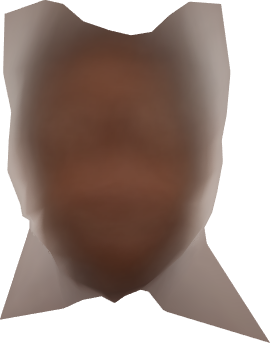}&
\includegraphics[height=2cm, clip=true]{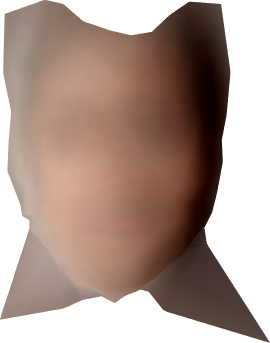}
\end{tabular}

}
\caption{Shape and appearance models learnt whilst training a StaTN on the dataset shown in Figure \ref{fig:fitting}.}
\label{fig:model}
\end{figure}

\section{Experimental Results}

We use the UMDFaces Dataset \cite{bansal2016umdfaces} in our experiments. We choose 750 identities with the highest number of images, comprising 61311 in total. This is a rather small dataset, however, we apply random cropping on images and batch normalisation to the convolutional layers as data augmentation.

We follow very similar architecture (5-6 convolutional layers with ReLU and pooling followed by a fully connected layer) for our localiser and recognition parts of the network.
We use 10 dimensions for both our statistical shape and appearance models. We trained our network with classification, texture and symmetry losses.
The learning rates of the localiser and recognition layers are 0.001 whereas shape and texture layers are 0.01 and 1, respectively.

In Figure \ref{fig:curve} we show the training and validation curves for our proposed StaTN network and an equivalent recognition network without spatial transformation. There is a modest but clear improvement in validation performance, even though our proposed network performs recognition with less information (since the grid is smaller than the image, part of the image data is discarded prior to recognition).

Figure \ref{fig:fitting} shows qualitative fitting results predicted by our network's grid generator. The sparse grid successfully locates the face even in images that are highly cluttered, noisy and badly cropped. Note that we trained our network without any supervision.

In Figure \ref{fig:model} we show the shape and appearance model learnt by our network. The shape model clearly resembles a face shape. Interestingly, the shape model does not appear to include the ears, but does sample a region of the shoulders and neck. The texture model has clearly interpretable principal components. The first two capture global lighting or skin colour changes, the third captures side to side lighting variation. The shape components are less easily interpretable but the second mode appearances to capture side to side 3D rotation of the face.

In Figure \ref{fig:meanfaces} we apply the StaTN to multiple images of the same person and then average the output of the bilinear sampler of our network. We show comparison between the raw average (1st row) and the sampled average (2nd row). The number of images for each subject is shown in parentheses. The averages of the resampled images are much sharper and more recognisable than the averages of the raw images. This shows that the StaTN is successfully establishing correspondence between the images.

\begin{figure}[!t] \centering
\noindent\resizebox{\textwidth}{!}{

\begin{tabular}{cccccc}

\includegraphics[height=2cm, clip=true]{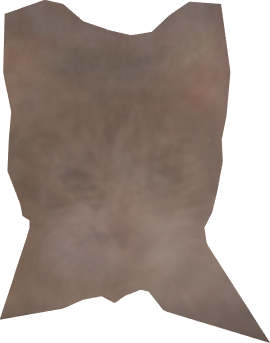}&
\includegraphics[height=2cm, clip=true]{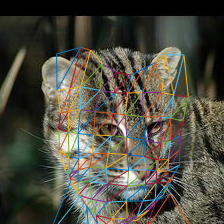}&
\includegraphics[height=2cm, clip=true]{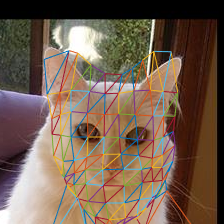}&
\includegraphics[height=2cm, clip=true]{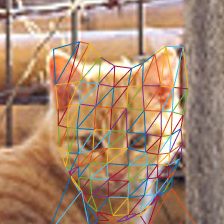}&
\includegraphics[height=2cm, clip=true]{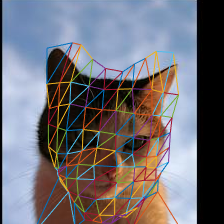}&
\includegraphics[height=2cm, clip=true]{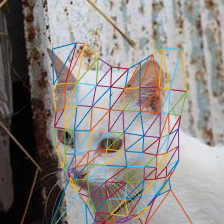}\\
\includegraphics[height=2cm, clip=true]{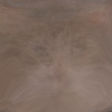}&
\includegraphics[height=2cm, clip=true]{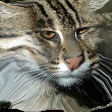}&
\includegraphics[height=2cm, clip=true]{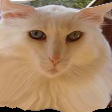}&
\includegraphics[height=2cm, clip=true]{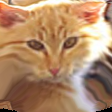}&
\includegraphics[height=2cm, clip=true]{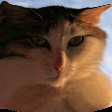}&
\includegraphics[height=2cm, clip=true]{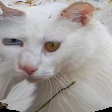}
\end{tabular}

}
\caption{Transfer learning on the CAT Dataset \cite{zhang2008cat}. The first image of the first row shows the shape model with the mean texture. The other images in the first row illustrate the sparse grid fitting results. The first image of the second row shows the mean texture only.The other images in the second row illustrate the output of the bilinear sampler of our network.}
\label{fig:cat}
\end{figure}

Finally, in Figure \ref{fig:cat} we show results for a completely unsupervised dataset. Here, we train on 10k images from the CAT Dataset \cite{zhang2008cat}. These images have no identity labels so we use only the texture and regularisation losses. We initialise with the face network trained in the previous experiment and finetune. Again, the network learns to consistently fit a meaningful grid to each image and constructs a plausible appearance model.

\newcommand{\averagesize}{2cm}
\begin{figure}[!t]
\centering
\noindent\resizebox{\textwidth}{!}{
\begin{tabular}{ccccccc}
\includegraphics[height=\averagesize, clip=true]{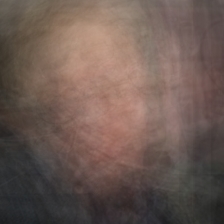}&
\includegraphics[height=\averagesize, clip=true]{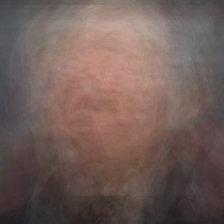}&
\includegraphics[height=\averagesize, clip=true]{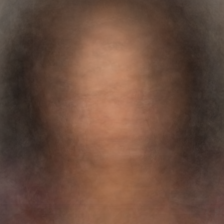}&
\includegraphics[height=\averagesize, clip=true]{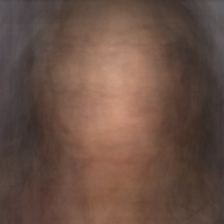}&
\includegraphics[height=\averagesize, clip=true]{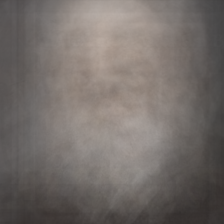}&
\includegraphics[height=\averagesize, clip=true]{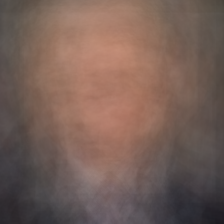}&
\includegraphics[height=\averagesize, clip=true]{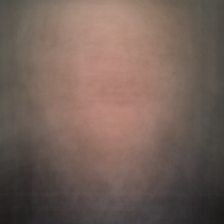}\\
\includegraphics[height=\averagesize, clip=true]{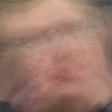}&
\includegraphics[height=\averagesize, clip=true]{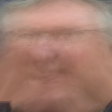}&
\includegraphics[height=\averagesize, clip=true]{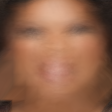}&
\includegraphics[height=\averagesize, clip=true]{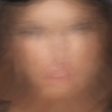}&
\includegraphics[height=\averagesize, clip=true]{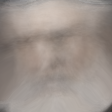}&
\includegraphics[height=\averagesize, clip=true]{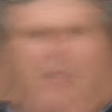}&
\includegraphics[height=\averagesize, clip=true]{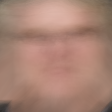}\\[-0.3em]

\tiny{S. Hawking(40)}&\tiny{Alex Ferguson(49)}&\tiny{Oprah Winfrey(71)}&\tiny{Adriana Lima(72)}&\tiny{A.Graham Bell(79)}&\tiny{G.W. Bush(106)}&\tiny{P.S. Hoffman(172)}
\end{tabular}
}
\caption{A set of averaged images per subject from the UMDFaces dataset \cite{bansal2016umdfaces}. 1st Row: Averaging raw face images of the same person. 2nd Row: Images that are obtained by applying the StaTN to multiple images of the same person. The number of images that are used for averaging is stated next to subject's name.}
\label{fig:meanfaces}
\end{figure}

\section{Conclusions}

We have presented a method that attempts to combine model and learning-based computer vision. By incorporating an explicit shape and appearance model along with a rigid transformation model, our StaTN network is able to explicitly learn dense, nonrigid correspondence. Moreover, the shape, appearance and pose parameters are interpretable and the shape and appearance models form components that can be reused in other networks or other settings. This reduces the ``black box'' nature of a CNN to some extent.

Using a StaTN as part of a network that is learning to solve a task, e.g. with a classification loss, then the network learns a notion of correspondence that is optimal for that task. This may be revealing about what information is most important for solving a particular task. When the texture loss is used in conjunction with learning a task, then the network learns to trade off sampling more of the image (and potentially sampling useful contextual information in the background) against attending to more easily compressible objects in the image. When the texture loss is used on its own, the network seeks the most compressible object class present in the training images.

The most obvious extension of our work would be to learn a 3D shape model and a 3D to 2D geometric projection model. This could be seen as extending work such as \cite{bas20173D,tewari2017mofa} by making the 3D model learnable. A 3D model allows 3D pose changes to be more efficiently modelled and occlusions to be dealt with explicitly, though this introduces a non-differentiable visibility function.

\bibliographystyle{splncs}

\end{document}